\documentclass[letterpaper]{article} 

\usepackage{aaai25}  
\usepackage{times}  
\usepackage{helvet}  
\usepackage{courier}  
\usepackage[hyphens]{url}  
\usepackage{graphicx} 
\usepackage{natbib}  
\usepackage{caption} 
\usepackage{multirow}

\graphicspath{{./images}}
\usepackage{amssymb}
\usepackage{algorithm}
\usepackage{algorithmic}
\usepackage{amsmath,amssymb,}
\usepackage{newfloat}
\usepackage{listings}
\DeclareCaptionStyle{ruled}{labelfont=normalfont,labelsep=colon,strut=off} 
\floatstyle{ruled}
\newfloat{listing}{tb}{lst}{}
\floatname{listing}{Listing}
\title{Evolvable Psychology Informed Neural Network for Memory Behavior Modeling}
    
\author{
    Xiaoxuan Shen\textsuperscript{\rm 1,\rm 2},
    Zhihai Hu\textsuperscript{\rm 1,\rm 2},
    Qirong Chen\textsuperscript{\rm 1,\rm 2},\\
    Shengyingjie Liu\textsuperscript{\rm 1,\rm 2},
    Ruxia Liang\textsuperscript{\rm 1,\rm 2},
    Jianwen Sun\textsuperscript{\rm 1,\rm  2}\equalcontrib
}
\affiliations {
    \textsuperscript{\rm 1}National Engineering Research Center of Educational Big Data, Central China Normal University, Wuhan 430079, China\\
    \textsuperscript{\rm 2}Faculty of Artificial Intelligence in Education, Central China Normal University, Wuhan 430079, China\\
    
}

\usepackage{bibentry}

\begin{document}

\maketitle

\begin{abstract}
Memory behavior modeling is a core issue in cognitive psychology and education. Classical psychological theories typically use memory equations to describe memory behavior, which exhibits insufficient accuracy and controversy, while data-driven memory modeling methods often require large amounts of training data and lack interpretability. Knowledge-informed neural network models have shown excellent performance in fields like physics, but there have been few attempts in the domain of behavior modeling. This paper proposed a psychology theory informed neural networks for memory behavior modeling named PsyINN, where it constructs a framework that combines neural network with differentiating sparse regression, achieving joint optimization. Specifically, to address the controversies and ambiguity of descriptors in memory equations, a descriptor evolution method based on differentiating operators is proposed to achieve precise characterization of descriptors and the evolution of memory theoretical equations. Additionally, a buffering mechanism for the sparse regression and a multi-module alternating iterative optimization method are proposed, effectively mitigating gradient instability and local optima issues. On four large-scale real-world memory behavior datasets, the proposed method surpasses the state-of-the-art methods in prediction accuracy. Ablation study demonstrates the effectiveness of the proposed refinements, and application experiments showcase its potential in inspiring psychological research.
\end{abstract}

%
\begin{links}
    \link{Code}{https://anonymous.4open.science/r/PsyINN-11FD}
    \link{Datasets}{https://anonymous.4open.science/r/PsyINN-11FD}
\end{links}

\section{INTRODUCTION}
Memory is an integral part of human cognitive processes. Developing models of memory behavior helps us better understand the essence and mechanisms of memory, which provides significant theoretical value to cognitive psychology, behavioral science, and neuroscience\cite{HistoryOverview}. Moreover, memory behavior modeling has important practical applications in educational, medical, and other real-world contexts.

Researchers typically use symbolic models (memory equations) to characterize human memory behavior, with the earliest memory equation dating back to Ebbinghaus's forgetting curve in 1885\cite{ebbinghaus}. Subsequently, particularly for scenarios involving multiple spaced repetitions in word learning, several new memory equations have been proposed, including ACT-R\cite{ACT-R} and MCM\cite{mcm}. However, there is currently no consensus in the theoretical community regarding the form of memory equations, and there are widespread controversies\cite{EquationControversy}. With the advancement of big data and artificial intelligence technologies. Many researchers have begun to explore data-driven approaches to memory behavior modeling\cite{flowerdew2015data}, such as Half-Life Regression (HLR)\cite{settles}. Deep learning techniques have also been introduced for learning behavior modeling\cite{tu2020deep} and memory behavior modeling\cite{ma,liu}. Although data-driven methods offer certain advantages in predictive performance, they still face challenges such as high demands for training data\cite{li2022interpretable}, the black-box nature of the prediction process\cite{rudin2022black}, and difficulties in ensuring the reliability of the predictive results\cite{wang2024survey}.

\begin{figure}[t]
\centering
\includegraphics[width=0.9\columnwidth]{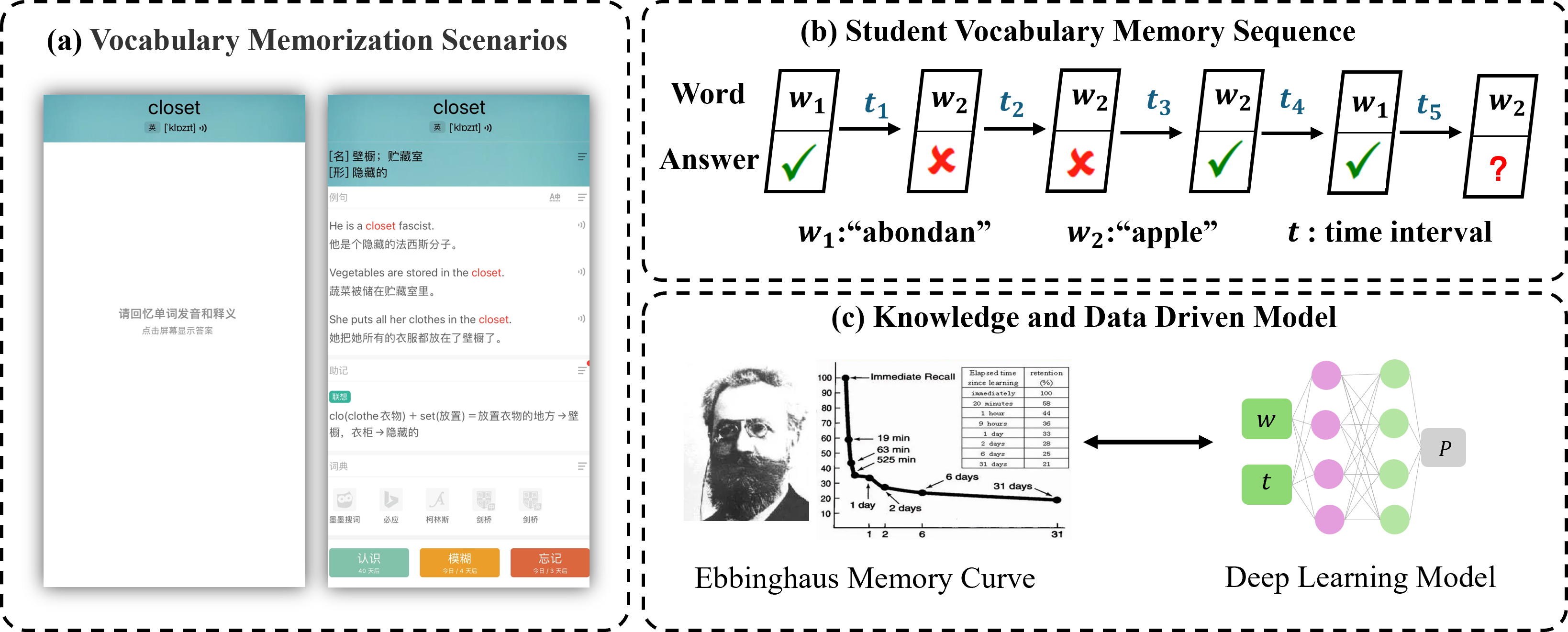} 
\caption{(a) shows vocabulary flashcards in MaiMemo. (b) provides an example of learners alternating between different vocabulary words. (c) illustrates our research motivation: using traditional theoretical equations to inspire neural networks, while neural networks can optimize traditional theoretical equations, achieving a synergistic effect}
\label{fig1}
\end{figure}

Knowledge-driven neural network models enhance the stability and interpretability of data-driven models by integrating domain knowledge, and they have achieved significant success in modeling natural processes, particularly in fields like physics\cite{karniadakis2021physics}. The Physics-Informed Neural Networks (PINNs)\cite{PINN} is the most representative method in this regard. It incorporates known physical equations and conditions to constrain the output of the neural network, thereby reducing the data requirements for model training and improving the stability of the output. This approach has led to substantial progress in solving problems such as Partial Differential Equations \cite{pdes} and turbulence calculations\cite{vortex_PDE_computatio}. However, knowledge-driven neural networks have been seldom explored in behavior modeling and cognitive modeling. This paper aims to combine memory theory equations with neural network models to build a more efficient and stable memory behavior model, as illustrated in Fig\ref{fig1}.However, unlike physical equations, memory equations often fail to fully explain memory behavior, leading to prediction errors and debates about their form. Additionally, there is widespread ambiguity in the descriptors used in memory equations. For example, memory strength, an important variable in memory equations, lacks clear definition and quantification methods. These challenges complicate the computation of memory theory equations.

To address these challenges in memory modeling, we propose a psychology theory-informed neural network for memory behavior modeling, named PsyINN. It employs a temporal neural network model for data-driven behavior modeling while utilizing symbolic regression to evolve symbolic memory equations, ultimately achieving joint optimization of the model. Specifically, first, to address the controversies in classical psychological theories and the ambiguity of descriptors, we propose a differential sparse regression-based method for the evolution of descriptors, enabling precise characterization of the descriptors and evolution of classical symbolic models. Secondly, to tackle gradient instability and local optimum issues in the joint optimization of neural networks and symbolic regression, we introduce a coefficient caching and alternating iterative optimization method. Ultimately, this approach aims to establish a more efficient, interpretable, and stable model for memory behavior.

The main contributions of this paper can be summarized as follows:
\begin{itemize}
\item To the best of our knowledge, this is the first work in the field of memory behavior modeling that attempts to integrate psychology theory into the construction of a knowledge-driven neural network. We propose PsyINN, which combines a temporal neural network model with a differential sparse regression framework, achieving joint optimization of the neural network and classical symbolic models.
\item To address the controversies in classical psychological theories and the ambiguity of descriptors, we propose a descriptor evolution method based on differential operators, enabling precise characterization of descriptors and evolution of classical symbolic models.
\item To tackle gradient instability and local optimum issues during model optimization, we introduce a coefficient caching mechanism and a multi-module alternating iterative optimization method.
\item We conducted extensive experiments on four large-scale real-world memory behavior datasets, demonstrating that the proposed method surpasses the current state-of-the-art memory modeling methods in predictive accuracy. Additionally, the application experiments show its potential for theoretical research.
\end{itemize}
\section{Related Work }

\subsection{Traditional Memory Theory Equations}
Traditional memory theory equations use mathematical models to describe the processes of memory and forgetting, aiming to predict the extent of an individual's memory retention at different points in time \cite{katkov2022mathematical}. The following are introductions to some classic traditional memory theory equations.

The earliest human memory model can be traced back to the approximate form of Ebbinghaus's forgetting curve \cite{ebbinghaus}, which describes the ratio of time saved during relearning to the time spent during the initial learning. 
Subsequently, some researchers have pointed out that the power law equation is more suitable than the exponential model for describing the forgetting process of memory \cite{wickelgren}. They suggest that the memory process can be represented by the generalized power law form.
Furthermore, some researchers have proposed the Adaptive Control of Thought-Rational (ACT-R) theory \cite{ACT-R} to describe and simulate human cognitive processes.  this theory is based on fundamental assumptions and principles of simulating cognitive activities, and it can explain and predict cognitive performance across various tasks and situations.With the development of big data technology and artificial intelligence, many researchers have begun to explore data-driven approaches to memory behavior modeling\cite{choffin}. Half-Life Regression (HLR) \cite{settles} is a spaced repetition model. This model combines psycholinguistic theory with modern machine learning techniques to indirectly estimate the "half-life" of words in students' long-term memory, and uses this half-life to characterize learners' memory properties. 

Traditional memory theoretical equations have played a significant role in memory research, but they also have some limitations, such as the inability to accurately quantify characteristic features, vague descriptions, and uncertain forms. Therefore, it is necessary to combine traditional memory theoretical equations with advanced machine learning techniques to enhance the accuracy and practicality of memory theoretical equations.

\subsection{Representation Learning Models}

Modeling student representation learning is a critical topic in educational assessment, learning systems, and long-term memory research. Its goal is to observe, present, and quantify students' knowledge states \cite{oludipe2023machine}. By modeling students' knowledge states, it is possible to estimate their mastery of different knowledge points and questions, thereby guiding teachers to effectively intervene and supplement students' lacking knowledge \cite{corbett1994knowledge}. The effectiveness of deep learning techniques in knowledge tracing has been demonstrated, leading to the development of a model called Deep Knowledge Tracing \cite{diao2023precise} to capture knowledge states from students' question-answer interaction sequences. In recent years, researchers have studied learners' knowledge states from various perspectives, including forgetting behavior \cite{chen2023enhanced}, question difficulty \cite{han2013understanding}, and review conditions \cite{shu2024improving}. Furthermore,The inherent "black box" nature of deep learning cannot help us accurately characterize the state of memory and its evolution process\cite{qamar2023understanding}. Therefore, optimizing deep learning-based learning representation models to better simulate and understand the process of memory behavior is a highly challenging task.

\subsection{Physics Informed Neural Networks}

In recent years, PINNs have demonstrated remarkable application potential and broad prospects in the fields of dynamics \cite{vortex_PDE_computatio}and fluid mechanics\cite{cai2021physics}. Unlike traditional data-driven neural networks, PINNs, as a machine learning model that combines deep learning with physical knowledge, incorporate physical laws into the neural network's loss function, allowing the neural network to make more accurate predictions and simulations while satisfying physical constraints. This enhances the model's physical interpretability and generalization ability. \cite{cuomo2022scientific}.In practical discrete memory systems, data often contains significant noise, and memory interaction points are more sparse, with precisely formulated memory theory equations often not existing within the memory system. Therefore, when applied to human memory behavior modeling, PINNs  struggle to perform effectively in the presence of substantial noise and inaccurately embedded knowledge\cite{inda2022physics}.

This paper proposes a psychological theory neural network for memory behavior modeling, named PsyINN. It constructs a framework that jointly optimizes neural networks with differential sparse regression, achieving precise representation of descriptors and the evolution of psychological symbolic models.

\begin{figure}[t]
\centering
\includegraphics[width=0.9\columnwidth]{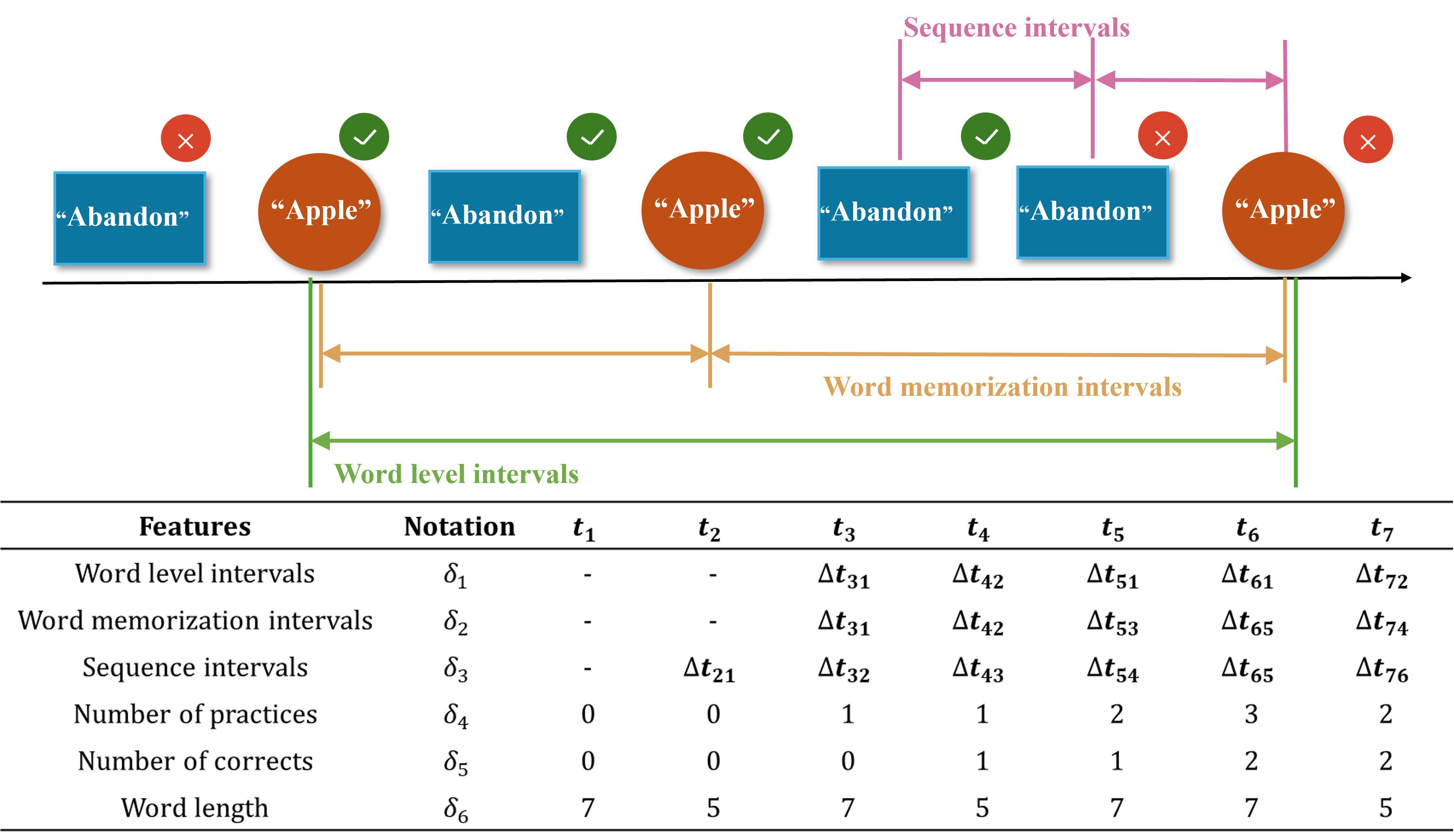} 
\caption{Input Feature Map}
\label{data}
\end{figure}

\section{Problem Statement}
Our goal is to predict the probability of a learner correctly answering any given word  based on their historical interaction data.
More specifically, for each learner $p$, we assume that a sequence of $m$ past interactions arranged in chronological order has been observed. The learner's word memory-related data is multivariate time series data $x_{{{p}_{1}}\to {{w}_{1}}:{{w}_{m}}}^{{{\text{t}}_{1:m}}}$ and $y_{{{p}_{1}}\to {{w}_{1}}:{{w}_{m}}}^{{{\text{t}}_{1:m}}}$.
The multivariate time series related to word $w_1$ for learner $p_1$ at the $m^{th}$ timestamp is denoted by $x_{{{p}_{1}}\to {{w}_{1}}}^{\text{t}m}$ , and the target time series at the $m^{th}$ timestamp is denoted by $y_{{{p}_{1}}\to {{w}_{1}}}^{{{\text{t}}_{m}}}$. 
We assume that each $x_{{{p}_{1}}\to wm}^{{{\text{t}}_{m}}}=[{{\delta }_{1}},{{\delta }_{2}},...,{{\delta }_{n}}]$  is a vector of length $n$. The definitions of ${{\delta }_{1}},{{\delta }_{2}},...,{{\delta }_{n}}$ are shown in Fig. \ref{data}. We would like to estimate the probability $y_{{{p}_{1}}\to {{w}_{1}}}^{{{\text{t}}_{m}}}$ of the student’s future performance on arbitrary word $x_{{{p}_{1}}\to wm}^{{{\text{t}}_{m}}}$.

\begin{figure*}[t]
\centering
\includegraphics[width=0.9\textwidth]{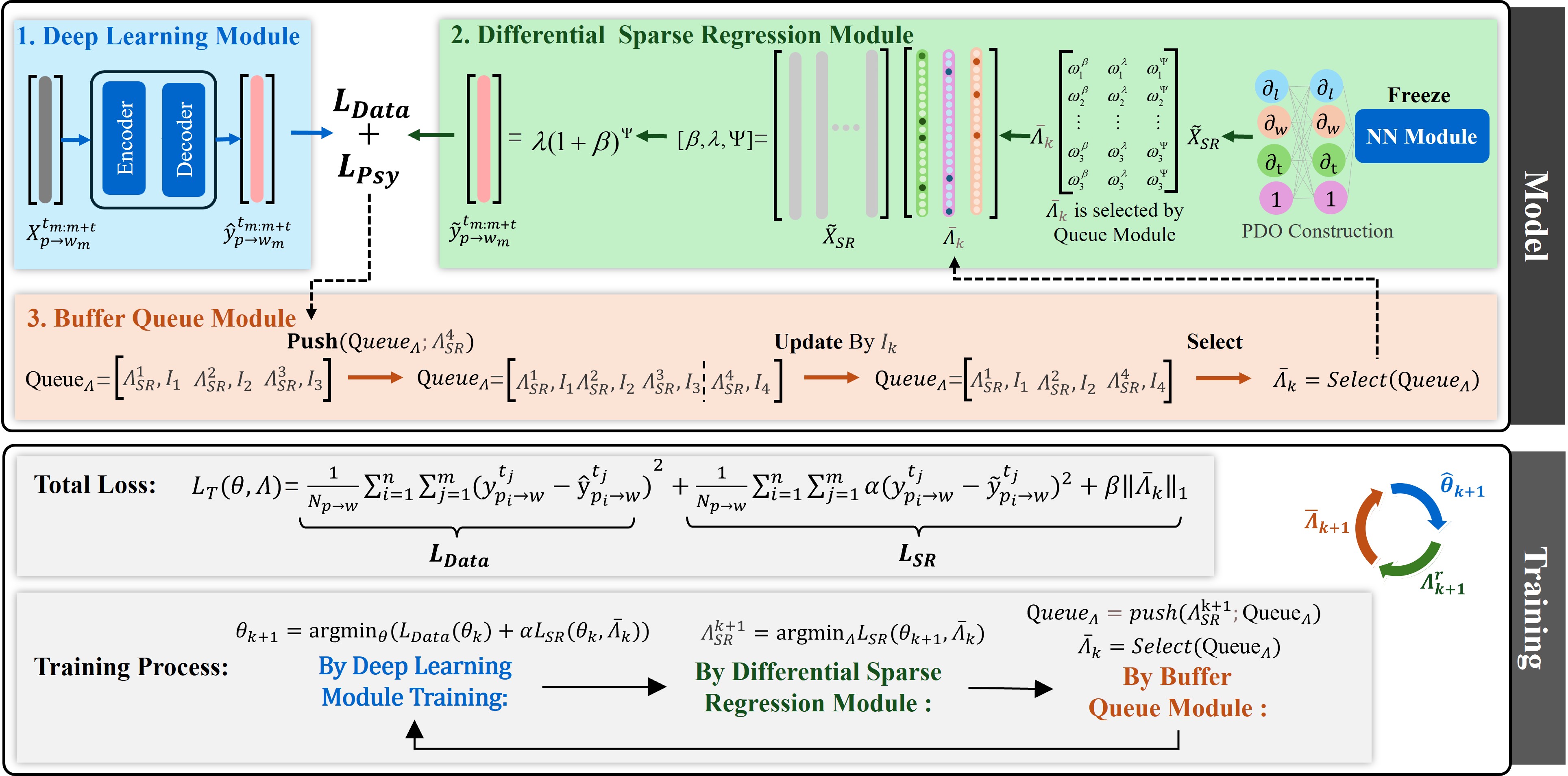} 
\caption{Model Structure Diagram}
\label{Model}
\end{figure*}


\section{Psychology-Inspired Neural Network}
We designed three different modules, namely the deep learning module, the Differential  Sparse Regression Module, and the buffer queue module, as shown in Fig\ref{Model}.
\begin{itemize}
\item \textbf{Deep Learning Module:} Models the learner's memory state to effectively predict memory retention probability. Additionally, its output serves as the input source for the Differential  Sparse Regression Module.
\item \textbf{Differential  Sparse Regression Module:} Extracts ordinary differential equations representing the learner's memory patterns. Furthermore, the obtained memory patterns can act as dynamic constraints during the training process of the deep learning module, thereby improving the accuracy and reliability of deep learning predictions.
\item \textbf{Buffer Queue Module:} To mitigate the gradient instability and local optima issues in differential sparse regression coefficients and enhance the diversity of theoretical equations, the differential sparse regression coefficients are  updated through a caching queue.
\end{itemize}
\subsection{Deep Learning Module}
Learning performance modeling can be used to predict a student's ability to respond to current questions. In the context of language learning, this method can predict a student's word proficiency at specific time points. Generally, a learner's word learning time series may include the following variables: time variables, reviewrelated variables, and word difficultyrelated variables (as shown in Fig.\ref{data}). Learning performance modeling can be predicted through deep learning networks, as shown in Equations \ref{Eq.10} and \ref{Eq.11}.
 \begin{gather}
        \tilde{h}_{{{p}_{1}}}^{{{\text{t}}_{m}}}=Encoder(x_{{{p}_{1}}\to {{w}_{1}}:{{w}_{m-1}}}^{{{\text{t}}_{1:m}}_{-1}},y_{{{p}_{1}}\to {{w}_{1}}:{{w}_{m-1}}}^{{{\text{t}}_{1:{{m}_{-1}}}}})  \label{Eq.10}\\
        \hat{y}_{{{p}_{1}}\to {{w}_{m}}}^{{{\text{t}}_{m}}}=Decoder(x_{{{p}_{1}}\to {{w}_{m}}}^{{{\text{t}}_{m-1}}},\tilde{h}_{{{p}_{1}}}^{{{\text{t}}_{m}}})\label{Eq.11}
\end{gather}
Here, Encoder denotes the encoder, and Decoder denotes the decoder. The encoder maps the historical memory behavior of learner $p_1$ to the memory state at time $t_m$, represented as ${\widetilde{h}}_{p_1}^{t_m}$ .The decoder predicts the recall probability $\hat{y}_{{{p}_{1}}\to {{w}_{m}}}^{{{\text{t}}_{m}}}$ of learner $p_1$  for word $w_{m}$ at time $t_{m}$  based on the learner's memory state and current memory behavior. In this study, DKT-F\cite{piech} is used as the encoder, and DNN is used as the decoder. Additionally, the deep learning module calculates the data loss by computing the MSE loss between the model output and the actual situation, with the specific calculation method given in Equation \ref{Eq.12}.
 \begin{gather}
        {{L}_{Data}}(\theta )=\frac{1}{{{N}}}\sum\limits_{i=1}^{n}{\sum\limits_{j=1}^{m}{{{(y_{{{p}_{i}}\to {{w}_{j}}}^{{{\text{t}}_{j}}}-\hat{y}_{{{p}_{i}}\to {{w}_{j}}}^{{{\text{t}}_{j}}})}^{2}}}} \label{Eq.12}
\end{gather}

\subsection{Differential  Sparse Regression Module}
To better reflect the subtle changes and trends in the learning process and enhance the descriptors' ability to describe learning progress, we introduce partial differential variables as features of the descriptors to more accurately depict learners' memory states at different time points. The deep learning module can accurately model learners' memory states and provides partial differential terms for describing the descriptors through automatic gradients. As shown in Equation \ref{Eq.13}, the final input for sparse regression is obtained by concatenating the original input $x_{{{p}_{1}}\to {{w}_{m}}}^{{{\text{t}}_{\text{m}}}}$ with its partial differential terms ${d\hat{y}_{{{p}_{1}}\to {{w}_{m}}}^{{{\text{t}}_{m}}}}/{dx_{{{p}_{1}}\to wm}^{{{\text{t}}_{m}}}}\;$.
 \begin{gather}
        {{\tilde{X}}_{SR}}=Concat([x_{{{p}_{1}}\to {{w}_{m}}}^{{{\text{t}}_{\text{m}}}},\frac{d\hat{y}_{{{p}_{1}}\to {{w}_{m}}}^{{{\text{t}}_{m}}}}{dx_{{{p}_{1}}\to wm}^{{{\text{t}}_{m}}}}]) \label{Eq.13}
\end{gather}
Here, $Concat(\cdot)$ denotes the concatenation function, and ${\widetilde{X}}_{SR}\in R^{2n\times z}$ represents the input items for sparse regression.

The differential sparse regression coefficient matrix (composed of $z$ descriptors) is denoted as $\mathrm{\Lambda}_{SR}\in R^{2n\times z}$. The calculation method for the differential sparse regression module is shown in Equation \ref{Eq.14}. Furthermore, we aim to retain only a few terms in the differential sparse regression model that capture the main trends in the data, while other negligible terms are considered insignificant\cite{guerra2021guide}. Therefore, the training objective of the differential sparse regression module is to optimize the Mean Squared Error (MSE) loss of the preset memory equation relative to the true values and the sparsity of the differential sparse regression coefficient matrix, thereby further constraining the deep learning model. The loss calculation method for the differential sparse regression module is shown in Equation \ref{Eq.15}.

 \begin{gather}
        \tilde{y}_{{{p}_{1}}\to {{w}_{m}}}^{{{\text{t}}_{m}}{{_{:}}_{m}}_{+\text{t}}}=f({{\tilde{X}}_{SR}}{{\Lambda }_{SR}})\label{Eq.14}
\end{gather}
\begin{equation} \label{Eq.15}
\small
\begin{split}
  {{L}_{SR}}(y,x;{{\theta }_{k}},{{\Lambda }_{k}}) &= \frac{1}{N}\sum\limits_{i=1}^{n}{\sum\limits_{j=\text{m+1}}^{\text{m+t}}{{{(y_{{{p}_{i}}\to {{w}_{j}}}^{{{\text{t}}_{j}}}-\tilde{y}_{{{p}_{i}}\to {{w}_{j}}}^{{{\text{t}}_{j}}})}^{2}}}}  \\ 
  &+ \lambda ||{{\Lambda }_{k}}|{{|}_{1}}
\end{split}
\end{equation}
Here, $f(\cdot)$ represents the form of traditional theoretical equations

\subsection{Buffer Queue Module}
Addressing the controversies and vagueness of descriptors in classical psychological theories, as well as alleviating gradient instability and local optimum issues in model optimization, we designed a caching mechanism for selecting differential sparse regression coefficient matrices. Specifically, during training, the coefficients learned by the   Differential Sparse Regression Module are continuously updated to the queue, and the better coefficients are selected to update the constraint equations, thereby better inspiring the neural network.We use the validation error of the sparse matrix as an importance factor to quantify the optimal sparse matrix coefficients and employ a buffer queue to store the top $k$ optimal sparse matrix coefficients. The calculation method for the importance factor is shown in the Equation \ref{Eq.16}.
 \begin{gather}
        {{I}_{\text{k}}}={{L}_{Data}}(y,x)+{{L}_{SR}}(y,{{X}_{SR}};\Lambda _{SR}^{k})\label{Eq.16}
\end{gather}
Where $\mathrm{\Lambda}_{SR}^k$ represents the coefficients of the $k^{th}$ sparse matrix, and $I_k$ represents the loss of the $k^{th}$ matrix in the sparse matrix set. Each time the sparse matrix coefficients are updated, the contents of the buffer queue are also updated accordingly. The specific process is shown in the equation \ref{Eq.17} and \ref{Eq.18}.
 \begin{gather}
        Queue{{e}_{\Lambda }}=push(\Lambda _{SR}^{k},{{I}_{k}}; Queue{{e}_{\Lambda }})\label{Eq.17}\\
        {{\bar{\Lambda }}_{k}}=Update(Queu{e}_{\Lambda })\label{Eq.18}
\end{gather}

Considering the importance factors in all cache queues and the inherent uncertainty of the sparse matrix, we propose three different strategies to generate the memory equation coefficients that constrain the predictive model: optimal selection replacement (selecting the optimal coefficient matrix according to the importance factor, as shown in Equation \ref{Eq.19}), random selection replacement (randomly selecting a coefficient matrix from the cache queue, as shown in Equation \ref{Eq.20}), and direct selection replacement (not using the cache queue, utilizing the currently obtained coefficient matrix in each iteration, as shown in Equation \ref{Eq.21}).
 \begin{gather}
        {{\bar{\Lambda }}_{k}}=Best\_Select(Queu{{e}_{\Lambda }})\label{Eq.19}\\
        {{\bar{\Lambda }}_{k}}=Random\text{ }\!\!\_\!\!\text{ }Select(Queu{{e}_{\Lambda }})\label{Eq.20}\\
        {{\bar{\Lambda }}_{k}}=\Lambda _{SR}^{k}\label{Eq.21}
\end{gather}
Here,${\bar{\mathrm{\Lambda}}}_k$ represents the equation coefficients constrained by the next round of the deep learning module. 

\subsection{Training Method}
To implement a descriptor evolution method based on differential operators, which achieves precise representation of descriptors, evolution of psychological symbolic models, and joint optimization of neural networks with classical symbolic models, we designed a training method to regulate the iteration of the three components. During the optimization of the deep learning module, the parameters of the descriptors in the  Differential  Sparse Regression Module are frozen. The deep learning module is optimized by calculating the MSE loss and equation loss between the model output and the actual situation. Equation \ref{Eq.22} provides the overall objective of the model, and Equation \ref{Eq.23} provides the parameter update method during model training.
 \begin{gather}
       {{\text{L}}_{T}}(\theta ,\Lambda )={{L}_{Data}}(\theta )+\alpha {{L}_{SR}}(\theta ,\Lambda )\label{Eq.22}\\
       {{\theta }_{k+1}}=\arg {{\min }_{\theta }}[{{L}_{Data}}({{\theta }_{k}})+\alpha {{L}_{SR}}({{\theta }_{k}},{{\bar{\Lambda }}_{k}})]\label{Eq.23}
\end{gather}
Here, $L_T$ is the overall objective of the model, and $\alpha$ represents the weight of the sparse regression formula loss. Similarly, when optimizing the Differential  Sparse Regression Module, the parameters of the deep learning module are frozen. The Differential  Sparse Regression Module is optimized by calculating the MSE loss between the model output and the actual situation, with the specific calculation method provided in Equation \ref{Eq.24}.
\begin{gather}
       \Lambda _{SR}^{k+1}=\arg {{\min }_{\Lambda }}{{L}_{SR}}({{\theta }_{k+1}},{{\bar{\Lambda }}_{k}})\label{Eq.24}
\end{gather}
After alternating iterations of the two modules, the coefficients $\mathrm{\Lambda}_{SR}^{k+1}$ of the  Differential Sparse Regression Module are pushed into the buffer queue module, and the buffer queue module then selects the coefficients to be used in the next round of the Differential  Sparse Regression Module, as shown in Equation \ref{Eq.17} and \ref{Eq.18}. The training method is described in Algorithm 1 (in Appendix).

\begin{table*}[t]
\centering
\setlength{\tabcolsep}{1mm}
\fontsize{9pt}{9pt}\selectfont 
\begin{tabular}{cllllllll}
\hline
 & \multicolumn{2}{c}{Duolingo} & \multicolumn{2}{c}{Duolingo-de} & \multicolumn{2}{c}{Duolingo-es} & \multicolumn{2}{c}{MaiMemo} \\ \cline{2-9} 
Model & \multicolumn{1}{c}{MAE} & \multicolumn{1}{c}{MAPE} & \multicolumn{1}{c}{MAE} & \multicolumn{1}{c}{MAPE} & \multicolumn{1}{c}{MAE} & \multicolumn{1}{c}{MAPE} & \multicolumn{1}{c}{MAE} & \multicolumn{1}{c}{MAPE} \\ \hline

HLR & 0.117±.002 & 13.788±.192 & 0.116±.001 & 13.371±.105 & 0.113±.001 & 13.269±.110 & 0.224±.001 & 22.887±.093 \\
Wickelgren & 0.117±.001  &  13.769±.074  &  0.115±.002  &  13.279±.231  &  0.113±.002  &  13.201±.194  &  0.227±.002  &  23.049±.174 \\
ACT-R & 0.116±.003   &   13.688±.279   &   0.117±.001   &   13.438±.100   &   0.113±.002   &   13.288±.195   &   0.228±.001   &   23.141±.071 \\
DAS3H & 0.118±.001   &   13.879±.112   &   0.123±.001   &   13.883±.135   &   0.115±.001   &   13.601±.033   &   0.222±.003   &   22.541±.315 \\ \hline
DKT & 0.116±.002   &   13.628±.252   &   0.115±.002   &   13.335±.234   &   0.112±.003   &   13.124±.264   &   0.231±.001   &   23.401±.086 \\
DKT-Forget     &     0.110±.002   &   13.039±.151   &   0.112±.002  &   13.011±.165  &   0.107±.003  &   12.656±.231   &   0.219±.002   &   22.351±.213 \\
SimpleKT   &  \underline{0.108±.003}   &   \underline{12.878±.273}   &   0.112±.003   &   12.972±.300   &   0.107±.004   &   12.634±.445   &   0.227±.003   &   23.104±.373 \\
QIKT   &   0.112±.001   &   13.136±.136   &   \underline{0.111±.001}   &   \underline{12.854±.125}   &   0.112±.001   &   12.874±.149   &   0.228±.002   &   23.276±.175 \\
FIFAKT   &  0.113±.002   &  13.379±.259   &  0.111±.002  &   12.879±.161   &   \underline{0.105±.003}  &   \underline{12.536±.275}  &   \underline{0.218±.006}  & \underline{22.112±.573} \\
MIKT   &   0.113±.004   &   13.326±.525   &   0.111±.003   &   12.892±.339   &   0.109±.004   &   12.778±.414  &   0.231±.004   &   23.433±.594  \\ \hline

PsyINN(HLR)   &   \textbf{0.105±.002*}   &   
\textbf{12.585±.016*}   &   0.108{±.002}   &   12.663{±.197}   &  \textbf{0.097{±.004}*}   &   \textbf{11.658{±.435}*}   &   0.214{±.009}   &   21.886{±.854} \\
PsyINN(Wick)   &   0.107{±.001}   &   12.742±.068  &   0.109{±.003}   &   12.692{±.285}   &   0.099{±.002}   &   11.846±.281   &  \textbf{0.210{±.003}*}  & \textbf{21.458{±.274}*}\\
PsyINN(ACT-R)  &   0.106{±.001}   &   12.619{±.099}   &  \textbf{0.102{±.004}*}&  \textbf{12.059{±.377}*}   &   0.099{±.002}   &   11.958{±.153}   &   0.211{±.003}   &   21.643±.376 \\ \hline

\end{tabular}
\caption{The overall prediction performance of all baseline models and PsyINN. The best results are highlighted in bold, and the best baseline model is underlined. The marker $*$ indicates p-value $<$ 0.05 in the significance test}
\label{table1}
\end{table*}
\begin{table*}[t]
\centering
\setlength{\tabcolsep}{1mm}
\fontsize{9pt}{9pt}\selectfont 
\begin{tabular}{cllllllll}
\hline
 & \multicolumn{2}{c}{Duolingo} & \multicolumn{2}{c}{Duolingo-de} & \multicolumn{2}{c}{Duolingo-es} & \multicolumn{2}{c}{MaiMemo} \\ \cline{2-9} 
Model & \multicolumn{1}{c}{MAE} & \multicolumn{1}{c}{MAPE} & \multicolumn{1}{c}{MAE} & \multicolumn{1}{c}{MAPE} & \multicolumn{1}{c}{MAE} & \multicolumn{1}{c}{MAPE} & \multicolumn{1}{c}{MAE} & \multicolumn{1}{c}{MAPE} \\ \hline
HLR & 0.117±.002 & 13.788±.192 & 0.116±.001 & 13.371±.105 & 0.113±.001 & 13.269±.110 & 0.224±.001 & 22.887±.093 \\
E-HLR &\textbf{0.111±.002*} & \textbf{13.191±.083*} & \textbf{0.112±.002*} & \textbf{13.019±.194*} & \textbf{0.108±.001*} & \textbf{12.786±.033*} & \textbf{0.221±.002*}& \textbf{22.479±.202*} \\ \hline

Wickelgren & 0.117±.001  &  13.769±.074  &  0.115±.002  &  13.279±.231  &  0.113±.002  &  13.201±.194  &  0.227±.002  &  23.049±.174 \\

E-Wickelgren & \textbf{0.106±.002*} & \textbf{12.688±.183*} & \textbf{0.105±.003*} & \textbf{12.318±.323*} & \textbf{0.103±.002*} & \textbf{12.230±.078*} & \textbf{0.210±.006*} & \textbf{21.428±.551*} \\ \hline
ACT-R & 0.116±.003   &   13.688±.279   &   0.117±.001   &   13.438±.100   &   0.113±.002   &   13.288±.195   &   0.228±.001   &   23.141±.071 \\
E-ACT-R& \textbf{0.106±.002*}& \textbf{12.711±.154*} & \textbf{0.110±.003*} & \textbf{12.777±.249*} & \textbf{0.105±.003*}& \textbf{12.476±.257*} & \textbf{0.212±.003*} & \textbf{21.724±.223*} \\ \hline

\end{tabular}

\caption{The overall performance of all traditional theoretical memory models and the memory models extracted by PsyINN. 'E' indicates the memory models extracted by PsyINN. The superior memory models are highlighted in bold. The marker $*$ indicates p-value $<$ 0.05 in the significance test}
\label{table2}
\end{table*}

\section{Experiment}
We employed four real-world datasets and constructed ten benchmark models, which are composed of two categories: Traditional Theoretical Equation Models and Representation Learning Models. For more details, refer to the Appendix.
\subsection{Results}

In this section, to demonstrate the effectiveness of our proposed PsyINN method and the ability of the optimized ordinary differential equations from  differential sparse regression to better describe learners' memory, we designed two comparative experiments: 1. The model comparison experiment, which compares the performance of the PsyINN method with all baseline models; 2. The theoretical equation comparison experiment, which compares the performance of the equations derived from the differential sparse regression module with traditional theoretical memory equations.

\textbf{Model Comparison Experiment:} Table \ref{table1} presents the performance comparison of the PsyINN neural network module with all baseline models, and we draw the following observations: (1) Compared with other baseline models, the neural network predictions obtained by PsyINN are significantly better than all traditional memory theoretical equation models and other neural network models on all datasets. (2) PsyINN is implemented based on DKT-F, but compared with DKT-F, the PsyINN model shows significant improvements on all datasets, indicating that memory theoretical equations can reflect the potential laws of memory to a certain extent, and the memory theoretical equations optimized by sparse regression can better guide the neural network. (3) The good performance of DKTF demonstrates the importance of time factors in long-term word memory. (4) The excellent performance of the SimpleKT\cite{ma}, QIKT\cite{liu}, and MIKT\cite{sun} models on the Duolingo dataset and its subsets proves that modeling learners' memory states around questions is effective. This also shows that the interrelationships between words have a significant impact on memory effects in complex word memory scenarios. (5) The good performance of FIFAKT\cite{nagatani} on all datasets demonstrates that both time and the inter-relationships between words jointly affect learners' memory states.

\textbf{Theoretical Equation Comparison Experiment: } Table \ref{table2} presents the performance comparison between the equations extracted by the PsyINN Differential  Sparse Regression Module and traditional theoretical memory equations. We draw the following observations: Compared with other traditional theoretical memory equations, the equations extracted by PsyINN are significantly better. This demonstrates the potential of PsyINN in uncovering memory patterns and highlights that incorporating partial differential terms in memory theoretical equations can better capture the subtle changes and trends in learners' memory states. 

\subsection{Ablation Experiment}
In this section, to demonstrate that our proposed caching mechanism for differential sparse regression coefficient matrices and the alternating iterative optimization method for multiple modules effectively alleviate gradient instability and local optimum issues in model optimization, we conducted ablation experiments analyzing the model components, cache queue selection strategies, and model iterative optimization methods.

\begin{figure}[t]
\centering
\includegraphics[width=0.9\columnwidth]{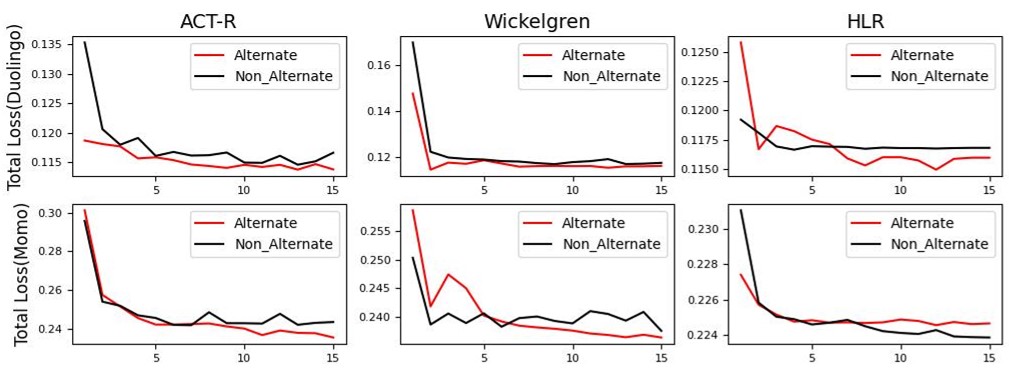} 
\caption{Training Error Trend of Alternating Gradient Descent Strategy on Duolingo and MaiMemo Datasets}
\label{alternate}
\end{figure}

\begin{figure}
\centering
\includegraphics[width=1\columnwidth]{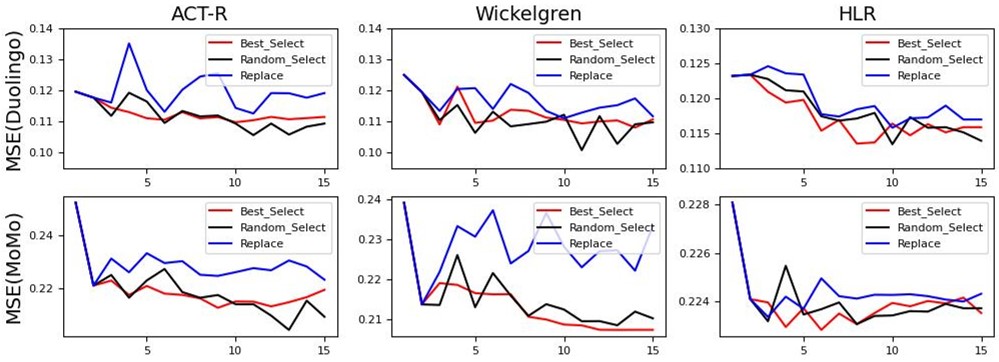} 
\caption{Change in Test Metrics with Different Selection Strategies on Duolingo and MaiMemo Datasets}
\label{select}
\end{figure}

\begin{table}[t]
\centering
\setlength{\tabcolsep}{1pt}
\fontsize{8pt}{8pt}\selectfont 
\begin{tabular}{cccccccccll}
\hline
\multirow{2}{*}{DKT-F} & \multirow{2}{*}{\begin{tabular}[c]{@{}c@{}}SR \end{tabular}} & \multirow{2}{*}{\begin{tabular}[c]{@{}c@{}}Select \end{tabular}} & \multicolumn{2}{c}{Duolingo} & \multicolumn{2}{c}{Duolingo-de} & \multicolumn{2}{c}{Duolingo-es} & \multicolumn{2}{c}{MaiMemo} \\ \cline{4-11} 
 &  &  & MAE & MAPE & MAE & MAPE & MAE & MAPE & \multicolumn{1}{c}{MAE} & \multicolumn{1}{c}{MAPE} \\ \hline
\checkmark &  &  & 0.109 & 13.012 & 0.1095 & 12.780 & 0.106 & 12.569 & 0.216 & 22.079 \\
 & \checkmark &  & 0.117 & 13.801 & 0.117 & 13.447 & 0.112 & 13.179 & 0.222 & 22.974 \\
\checkmark & \checkmark &  & \underline{0.105} & \underline{12.661} & \underline{0.112} & \underline{13.008} & \underline{0.105} & \underline{12.474} & \underline{0.206} & \underline{21.078} \\
 & \checkmark & \checkmark & 0.116 & 13.708 & 0.114 & 13.256 & 0.112 & 13.241 & 0.223 & 22.744 \\
\checkmark & \checkmark & \checkmark & \textbf{0.102} & \textbf{12.362} & \textbf{0.108} & \textbf{12.753} & \textbf{0.093} & \textbf{11.249} & \textbf{0.197} & \textbf{20.195} \\ \hline
\end{tabular}
\caption{Module Ablation Experiment.The highest results are highlighted in bold, and the second-best results are underlined}
\label{table3}
\end{table}

\textbf{Model Ablation Result: }To demonstrate the feasibility and effectiveness of PsyINN, we explore the impact of various modules on PsyINN. Specifically, we conduct variant experiments by selecting one or two modules from the neural network module, differential sparse regression module, and buffer queue module. Additionally, we choose the HLR memory equation as the preset equation for the differential sparse regression module and use the random selection replacement strategy as the buffer queue selection strategy. From Table \ref{table3}, we draw the following observations: (1) Compared to the DKT-F model, the models with the inclusion of traditional theoretical equation constraints show significant improvements, indicating that traditional theoretical equations can inspire neural network training. (2) Compared to the SR model, the SR model with the selection strategy shows no difference in performance, suggesting that the selection strategy is essential for optimizing the instability of partial differential term inputs in the differential sparse regression module. (3) Compared to the PsyINN model without the selection strategy, the PsyINN model with the selection strategy consistently exhibits significant advantages, indicating that choosing an appropriate strategy effectively alleviates gradient instability and local optima issues.


\textbf{Buffer Queue Selection Strategies: } To explore the impact of different selection strategies on model performance, we validated the three proposed selection strategies: direct selection replacement, optimal selection replacement, and random selection replacement. Fig. \ref{select}  shows the ablation experiment results for different selection strategies, where each curve represents the average of five test metrics. Black represents the random selection strategy, red represents the optimal selection replacement, and blue represents the direct selection replacement. We draw the following conclusions: (1) Models using selection replacement strategies significantly outperform those without, indicating that appropriate replacement strategies can effectively mitigate gradient instability. (2) Random selection replacement generally performs better than optimal selection replacement, suggesting that introducing a degree of randomness helps the model escape local optima.

\textbf{Alternating Iteration Strategy:} To explore the impact of the alternating gradient descent algorithm on training, we specifically compared two strategies: alternating iteration and non-alternating iteration during the training process of the neural network module and the differential sparse regression module. Fig. \ref{alternate} shows the ablation experiment results for different alternating strategies, where each curve represents the average of five training loss values. The black curve represents non-alternating iteration, while the red curve represents alternating iteration. It is evident that, in most cases, the training loss with alternating iteration outperforms that without alternating training. We hypothesize that the alternating gradient descent strategy helps the model adjust the parameter optimization direction more effectively, thereby finding the global optimum more efficiently and avoiding local optima.
\begin{figure}[t]
\centering
\includegraphics[width=1\columnwidth]{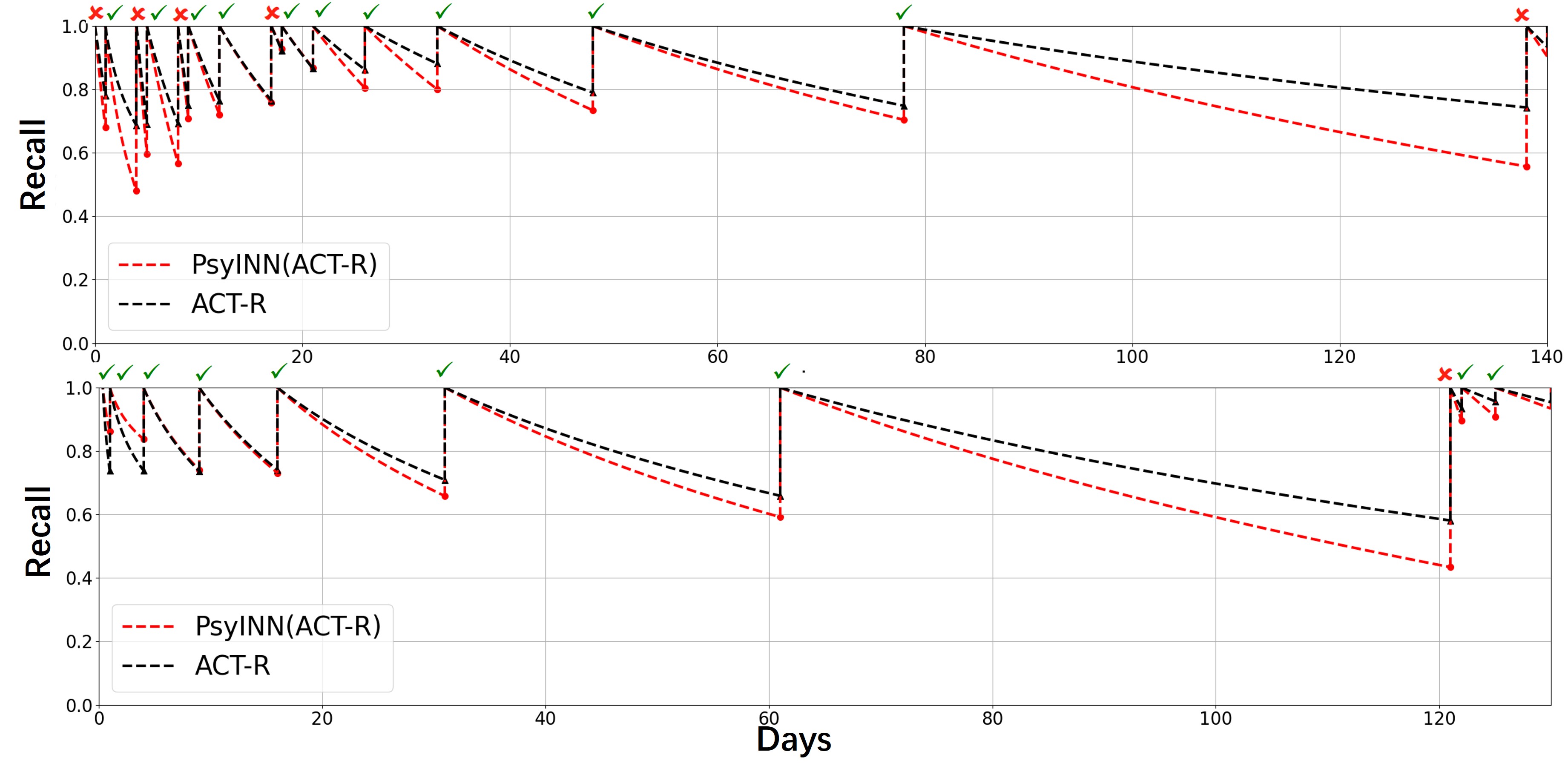} 
\caption{K-study Graph of Model Predicting Students' Word Learning Trajectory}
\label{k-day}
\end{figure}
The following images support the conclusions discussed in the "Description of Equation Values" section in the Application chapter.
\begin{figure}[t]
\centering
\includegraphics[width=1\columnwidth]{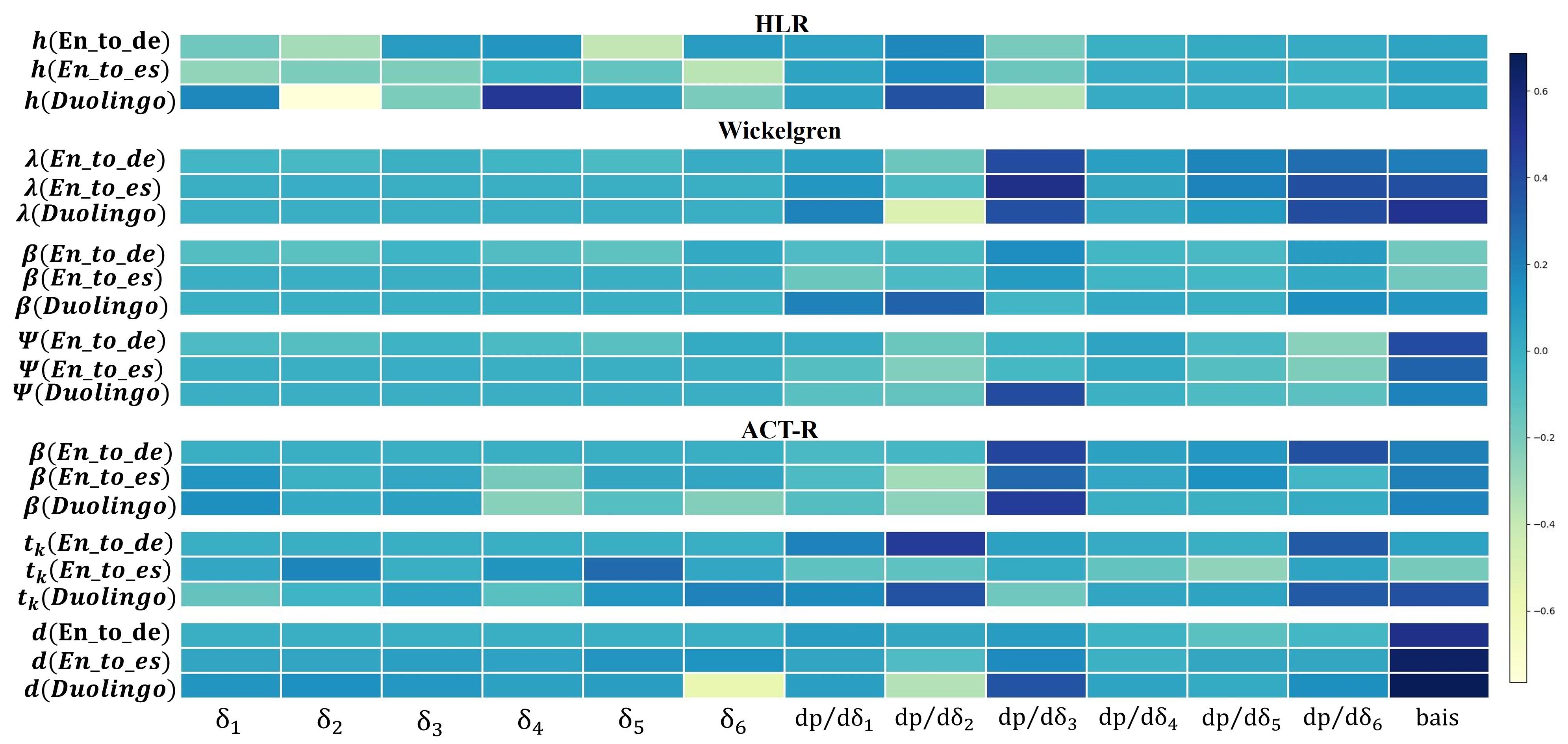} 
\caption{Heatmap of Descriptor Weights}
\label{heatmap}
\end{figure}
\section{Applications}
\textbf{K-study graph:} 
In the process of language acquisition, repeated practice ensures the effectiveness of learning, and designing review intervals to maximize memory retention is a crucial aspect of language learning. Fig. \ref{k-day}  shows the model comparison results of user data under two different learning conditions: effective learning and fluctuating learning outcomes. PsyINN can better predict the probability of a learner recalling a word at a specific point in time, thereby helping learners memorize words more effectively.

\textbf{Description of Equation Values:} In psychology, the descriptors of traditional memory equations are often vague and uncertain, and the exact form of these memory equations is highly debated. From the observations in Fig.\ref{heatmap}, we draw the following conclusions: (1) The parameter weights of descriptors in the memory ordinary differential equations show a greater tendency towards the partial differential terms, indicating that our introduction of partial differential terms in the memory equation can better model the learners' memory states, potentially clarifying the field of memory modeling and identifying key factors that influence memory. (2) The different target languages lead to variations in equation parameters, suggesting that different learning materials have an impact on the overall memory process.

\section{Conclusion}
This paper proposes a psychological theory neural network for memory behavior modeling to address the vagueness and uncertainty in traditional memory theory equations, achieving joint optimization of neural networks with classical symbolic models. 
On four large-scale real-world memory behavior datasets, our proposed PsyINN model outperforms state-of-the-art methods. At the same time, it enriches traditional memory theory equations into memory partial differential equations, achieving precise representation of descriptors and the evolution of psychological symbolic models, providing a novel research method for the current trend of discovering social science laws. However, our work has certain limitations. This study adopts a fixed equation form, which may limit the model's flexibility and adaptability to different learning processes. 

\section{Acknowledgments}
This work was financially supported by the National Science and Technology Major Project (2022ZD0117103), National Natural Science Foundation of China (62293554), Higher Education Science Research Program of China Association of Higher Education (23XXK0301), Hubei Provincial Natural Science Foundation of China (2022CFB414, 2023AFA020), and Fundamental Research Funds for the Central Universities (CCNU24AI016).
\bibliography{aaai25}


\appendix
\section{Appendix}
In this appendix, we present the training methods, experimental setup, hyperparameter details, and benchmark models.

The following algorithm is the "Training Method" from the psychology-inspired neural network section.
\begin{algorithm}[H]
\caption{PsyINN}
\label{alg:algorithm}
\textbf{Input}: student's word learning time series include time, review, and word difficulty, target data $Y$, number of epoch $N$,the initial sparse matrices set $\mathrm{\Lambda}_{SR}^k$\\
\textbf{Output}: Well-trained Parameters of the sparse matrices and model
\begin{algorithmic}[1] 
\STATE Let  ${\bar{\mathrm{\Lambda}}}_k=\mathrm{\Lambda}_{SR}^0$
\WHILE{condition}
\STATE Fix Sparse Matrix\ $\mathrm{\Lambda}_{SR}^k$
\FOR{$i=1:N$} 
\STATE Train Parameters of Deep Learning Module
\ENDFOR
\STATE Fix Parameters of Deep Learning Model
\FOR{$i=1:N$} 
\STATE Train Differential  Sparse Regression Module
\ENDFOR
\STATE Push $\mathrm{\Lambda}_k$ into the buffer queue and calculate the importance factors $I_k$ for all $\mathrm{\Lambda}_k^n$
\IF {$n > n_{max}$}
\STATE Remove Sparse Matrix with the Lowest Importance Factor
\ENDIF
\STATE Obtain New Coefficient Matrix According to Equation \ref{Eq.18}
\ENDWHILE
\STATE \textbf{return} solution
\end{algorithmic}
\end{algorithm}

\section{Experiment }
\subsection{Datasets}
We evaluated the performance of PsyINN on 4 publicly available commonly used datasets: Duolingo,Duolingo-de,Duolingo-es and MaiMemo:
\begin{itemize}
\item \textbf{Duolingo:} This dataset is a real-world dataset widely used in language learning applications. It contains language learning logs of 115,222 learners of English, French, German, Italian, Spanish, and Portuguese, recording 12.8 million student word courses and practice course logs.
\item \textbf{Duolingo-de:} From the Duolingo dataset, we extracted user data of those learning German using the English UI interface and named it the Duolingo-de subset. This subset contains 117,037 log data points from 2,485 learners.
\item \textbf{Duolingo-es:} From the Duolingo dataset, we extracted user data of those learning Spanish using the English UI interface and named it the Duolingo-es subset. This subset contains 271,854 log data points from 5,706 learners.
\item \textbf{MaiMemo:} This dataset comes from China’s most popular English learning application “MaiMemo”. It contains 200 million user learning records and 17,081 English words.
\end{itemize}
\subsection{Baselines}
To evaluate the effectiveness and robustness of our proposed PsyINN model, we compare it with the following state-of-the-art deep learning models and traditional theoretical models.

\subsubsection{Traditional Theoretical Equation Models}
\begin{itemize}
\item \textbf{HLR:} A traditional theoretical regression model that considers both forgetting and lexical features.\cite{settles}
\item \textbf{Wickelgren:} A traditional theoretical regression model based on the generalized power law memory model theory.\cite{wickelgren}
\item \textbf{ACT-R:} A traditional theoretical regression model based on the Adaptive Character of Thought–Rational \cite{ACT-R}
\item \textbf{DAS3H:} A model that estimates the parameters of the generalized power law equation using big data, enabling regression predictions of learners' memory states. \cite{choffin}
\end{itemize}

\subsubsection{Representation Learning Models}

\begin{itemize}
\item \textbf{DKT:} A deep learning model based on LSTM that models learners' knowledge states and predicts their future knowledge states. \cite{piech}
\item \textbf{DKT-Forget :} An improved version of DKT that incorporates students' forgetting behaviors.  \cite{piech}
\item \textbf{FIFAKT:} An improved version of DKT that combines forgetting behaviors with an attention mechanism. \cite{nagatani}
\item \textbf{SimpleKT:} An improved version of DKT that incorporates individual differences in problems. \cite{ma}
\item \textbf{QIKT:} An improved version of DKT that combines problem difficulty with learners' knowledge states.  \cite{liu}
\item\textbf{MIKT:} An improved version of DKT that combines learners' domain knowledge states with conceptual knowledge states. \cite{sun}
\end{itemize}

\subsection{Experimental Setup}
 To train and validate the models, we used 80\% of the student sequences, reserving the remaining 20\% for model evaluation. All models were trained using the Adam optimizer for 200 epochs, with each model trained five times. We applied an early stopping strategy, halting optimization if the MAE score on the validation set did not improve over the last 10 epochs. The DKT-F module was used as the neural network component. We searched for the optimal learning rate and embedding dimension $d$ within the ranges of [2e-3, 3e-3, 4e-3, 5e-3, 6e-3] and [64, 256], respectively. All models were implemented in PyTorch and trained on a cluster of Linux servers equipped with NVIDIA GeForce RTX 3060 GPUs. Given the inconsistencies in evaluation standards between the Duolingo and MaiMemo datasets, we primarily used Mean Absolute Percentage Error (MAPE) as the main evaluation metric, with Mean Absolute Error (MAE) as a secondary metric. The evaluation metrics are calculated as follows.
\end{document}